\documentclass[letterpaper, 10 pt, journal, twoside]{ieeetran}

\IEEEoverridecommandlockouts                              

\usepackage{amsmath,amsfonts,amssymb}
\usepackage[dvipsnames]{xcolor}
\usepackage{algorithmic}
\usepackage{algorithm}
\usepackage{array}
\usepackage[labelfont=bf,margin=0.25in,font=small]{caption}
\usepackage{subcaption}
\usepackage{textcomp}
\usepackage{stfloats}
\usepackage{url}
\usepackage{verbatim}
\usepackage{graphicx}
\usepackage{cite}
\usepackage{nicematrix}
\usepackage{float}
\usepackage{tikz}
\usepackage{color}

\usepackage{multicol}
\usepackage{multirow}
\usepackage{pifont}
\newcommand{\cmark}{\ding{51}}
\newcommand{\xmark}{\ding{55}}
\usepackage{booktabs}
\usepackage{array}
\usepackage[most]{tcolorbox}
\usepackage{scalerel}
\usepackage{mathtools}
\usepackage{soul}
\usepackage[colorlinks]{hyperref}

\definecolor{steelblue}{HTML}{8CC9F7}
\definecolor{green}{HTML}{8AEEA8}

\newtcolorbox{mybox}{
  enhanced,
  colback=steelblue!50!white,
  colframe=steelblue,
  fonttitle=\bfseries,
  boxrule=0pt,  
  drop shadow
}

\newtheorem{example}{Example}

\newtheorem{definition}{Definition}

\newcommand{\Always}{\square}
\newcommand{\Eventually}{\lozenge}
\newcommand{\Until}{\mathcal{U}}

\newcommand{\signal}{\mathbf{s}}
\newcommand{\unrolled}{\mathcal{S}}
\newcommand{\masked}{\mathcal{M}}

\newcommand{\stlcgpp}{\texttt{STLCG++}}
\newcommand{\stlcg}{\texttt{STLCG}}

\newcommand{\trajectory}{\mathbf{x}}
\newcommand{\controls}{\mathbf{u}}
\newcommand{\robustnesstrace}{\boldsymbol{\tau}}

\newcommand{\logsumexp}{\mathrm{logsumexp}}

\newcommand{\softmin}{\mathrm{softmin}}
\newcommand{\softmax}{\mathrm{softmax}}
\newcommand{\smoothmask}{\widetilde{\mathrm{mask}}}

 \newcommand*\rounded[2]{\tikz[baseline=(char.base)]{
        \node[shape=circle,inner sep=2pt,fill=#2] (char) {\textcolor{#2}{#1}}}}

\title{STLCG++: A Masking Approach for Differentiable Signal Temporal Logic Specification
}

\author{Parv Kapoor$^{1,2}$, Kazuki Mizuta$^{3}$, Eunsuk Kang$^{1}$ and Karen Leung$^{3,4}$
\thanks{Manuscript received: February 4th, 2025; Revised May 11th, 2025; Accepted June 23rd, 2025.}%
\thanks{This paper was recommended for publication by Editor Lucia Pallottino upon evaluation of the Associate Editor and Reviewers' comments.
PK is supported by a National Science Foundation award under Grant No. 2144860. KM is supported by the UW+Amazon Science Hub Fellowship and Nakajima Foundation Fellowship.} 
\thanks{$^{1}$Parv Kapoor and Eunsuk Kang are with School of Computer Science, Carnegie Mellon University
        {\tt\footnotesize \{parvk,eunsukk\}@andrew.cmu.edu}}%
\thanks{$^{2}$Parv Kapoor is also affiliated with General Robotics}%
\thanks{$^{3}$Kazuki Mizuta and Karen Leung are with University of Washington,
        {\tt\footnotesize \{mizuta,kymleung\}@uw.edu}}%
\thanks{$^{4}$Karen Leung is also affiliated with NVIDIA}%
\thanks{Digital Object Identifier (DOI): 10.1109/LRA.2025.3588389}
}

\renewcommand{\baselinestretch}{1.0}

\begin{document}

\maketitle

\begin{abstract}
Signal Temporal Logic (STL) offers a concise yet expressive framework for specifying and reasoning about spatio-temporal behaviors of robotic systems. Attractively, STL admits the notion of \textit{robustness}, the degree to which an input signal satisfies or violates an STL specification, thus providing a nuanced evaluation of system performance. In particular, the differentiability of STL robustness enables direct integration to robotic workflows that rely on gradient-based optimization, such as trajectory optimization and deep learning. However, existing approaches to evaluating and differentiating STL robustness rely on recurrent computations, which become inefficient with longer sequences, limiting their use in time-sensitive applications. In this paper, we present \stlcgpp, a masking-based approach that parallelizes STL robustness evaluation and backpropagation across timesteps, achieving significant speed-ups compared to a recurrent approach.
We also introduce a smoothing technique to enable the differentiation of \textit{time interval bounds}, thereby expanding STL's applicability in gradient-based optimization tasks involving spatial and temporal variables. Finally, we demonstrate \stlcgpp's benefits through three robotics use cases and provide JAX and PyTorch libraries for seamless integration into modern robotics workflows. Project website with demo and code: \url{https://uw-ctrl.github.io/stlcg/}.
\end{abstract}

\vspace{-7mm}
\section{Introduction}

\IEEEPARstart{M}{any} robot planning tasks hinge on meeting desired spatio-temporal requirements, like a quadrotor navigating specific regions within strict time windows.
Signal temporal logic (STL) \cite{MalerNickovic2004} presents an attractive formalism to describe spatio-temporal specifications as it is designed to operate over \textit{real-valued} time-series input rather than discrete propositions.
In particular, STL is equipped with \textit{quantitative} semantics, or \textit{robustness formulas}, which measure how well a given robot trajectory satisfies a requirement.
With some smoothing approximations in place, it becomes efficient and stable to \textit{differentiate} STL robustness within gradient-based optimization methods---the key to many robot control and learning applications. As such, we have seen a growing interest in the inclusion of STL objectives/constraints in various optimization-based robotics problems utilizing gradient descent as a solution method, such as trajectory optimization \cite{PantAbbasEtAl2017}, deep learning \cite{MengFan2024}, and control synthesis \cite{LiuMehdipourEtAl2021}.

Recently, \stlcg \cite{LeungArechigaEtAl2021} was introduced as a general framework for encoding any STL robustness formula as a computation graph, leveraging modern automatic differentiation (AD) libraries for evaluation and backpropagation.
The \stlcg\ (PyTorch) library democratized STL for robotics and deep learning communities, enabling recent work in gradient-based optimization with STL objectives/constraints
\cite{ZhongRempeEtAl2023,LeungPavone2022,VeerLeungEtAl2023,DeCastroLeungEtAl2020}.


\begin{figure}
    \centering
    \captionsetup{width=\linewidth}
    \includegraphics[width=\columnwidth]{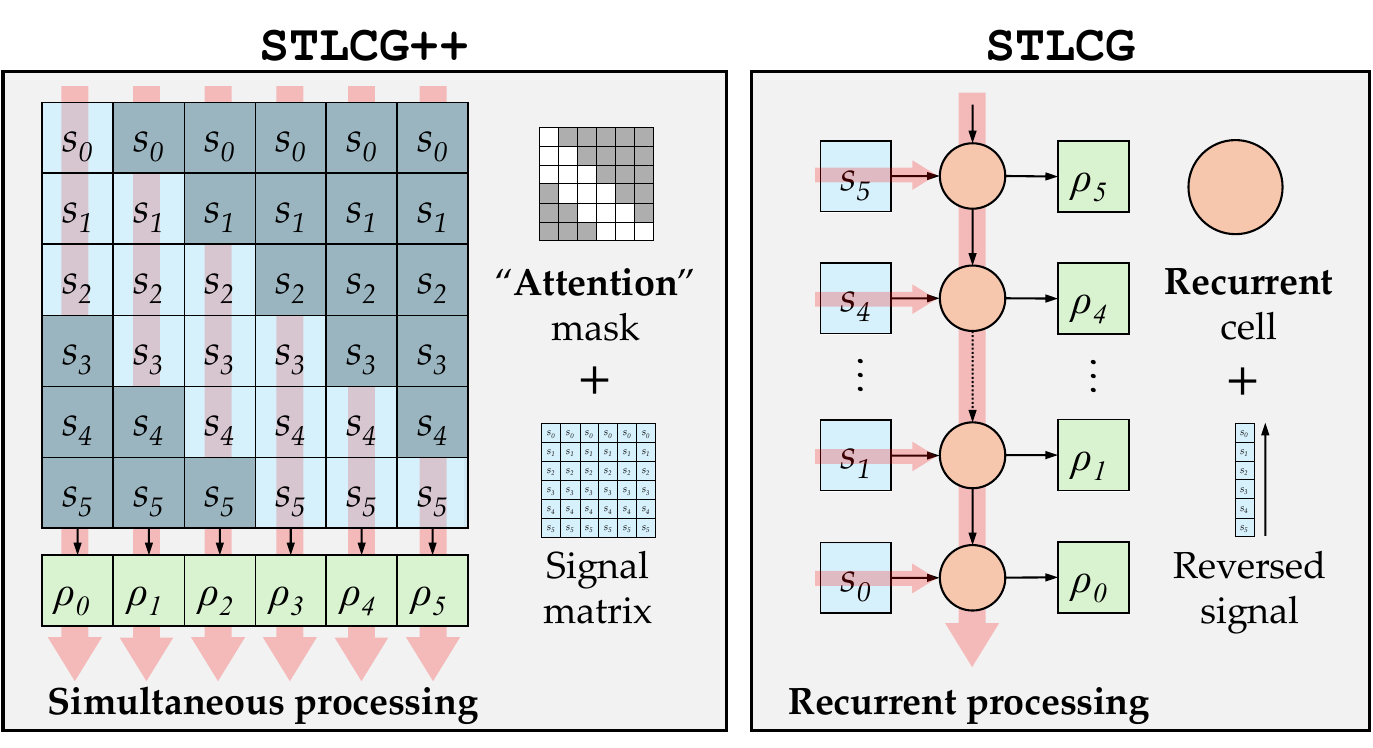}
    \caption{We propose \stlcgpp, a masking approach to evaluating and backpropagating through STL robustness formulas. The masking approach offers stronger computational, theoretical, and practical benefits compared to \stlcg, a recurrent approach.}
    \label{fig:hero}
    \vspace{-5mm}
\end{figure}

To construct the computation graph for any STL robustness formula, \stlcg\ processes the time-series input \textit{recurrently} (see Fig.~\ref{fig:hero} right), primarily inspired by how recurrent neural networks (RNNs) \cite{HochreiterSchmidhuber1997} process sequential data.
While consistent with the semantics of STL robustness, recurrent processing leads to the forward and backward passes being comparatively slower than other non-recurrent operations---a widely observed drawback of RNNs. These sequential operations limit \stlcg's capability for efficiently handling long sequence lengths in offline and online settings, especially when combined with other demanding computations, e.g., running foundation models.
More recently, attention-based neural architectures, such as transformers \cite{VaswaniShazeerEtAl2017}, have demonstrated superior performance in processing sequential data, particularly on GPU hardware. The key to the transformer architecture is the self-attention operation, which operates on all input values \textit{simultaneously} rather than recurrently.

Inspired by the masking mechanism in transformer architectures, we present \stlcgpp, a masking approach to evaluate and backpropagate through STL robustness for long sequences more efficiently than \stlcg, a recurrent-based approach (see Fig.~\ref{fig:hero} left).
\stlcgpp\ opens new possibilities for using STL requirements in long-sequence contexts, especially for online computations, paving the way for further advancements in spatio-temporal behavior generation, control synthesis, and analysis for robotics applications.

\noindent\textbf{Contributions.} The contributions of this paper are fourfold.
\textbf{(i)} We present \stlcgpp, a masking-based approach to computing STL robustness, and demonstrate the computational, mathematical, and practical benefits over \stlcg, a recurrent-based approach.
\textbf{(ii)} We introduce a smoothing function on the time interval of temporal operators and, with the proposed masking approach, enable the differentiability of STL robustness values with respect to time interval parameters.We apply a smooth masking mechanism to enable differentiation with respect to time interval parameters. To the best of the authors' knowledge, this is the first application of differentiating with respect to time parameters of STL formulas.
\textbf{(iii)} We demonstrate the benefits of our proposed masking approaches with several robotics-related problems ranging from unsupervised learning, trajectory planning, and deep generative modeling.
\textbf{(iv)} We provide \textit{two} open-source \stlcgpp\ libraries, one in JAX and another in PyTorch, and demonstrate their usage via the examples studied in this paper.
JAX and PyTorch are well-supported Python libraries used extensively by the deep learning and optimization communities.

\section{Related work}

As STL has been used for various applications, a variety of STL libraries across different programming languages have been developed, including Python, C++, Rust, and Matlab.
Given that Python is commonly used for robotics research, Tab.~\ref{tab:stltools} compares recent STL Python packages regarding automatic differentiation (AD), vectorization, and GPU compatibility.
Most libraries offer evaluation capabilities of a single signal, or their design is tailored towards a specific use case, making it difficult to extend or apply them to new settings. If users want to perform an optimization utilizing STL robustness formulas, a separate optimization package (e.g., CVXPY \cite{DiamondBoyd2016}, Drake \cite{Tedrakeothers2019}) is often required.

RTAMT \cite{YamaguchiHoxhaEtAl2023} was introduced as a unified tool for offline and online STL monitoring with an efficient C++ backend. It has received widespread support and has superseded other alternatives in terms of usage. However, RTAMT performs CPU-based signal evaluation and lacks differentiation and vectorization capabilities, limiting its efficiency in handling and optimizing over large datasets, where AD and GPU compatibility are crucial.
\stlcg\ was the first to introduce vectorized STL evaluation and backpropagation by leveraging modern AD libraries. However, \stlcg\ faced scalability challenges due to its underlying recurrent computation.
\stlcgpp\ addresses these scalability limitations by eliminating recurrent operations and optimizing GPU capabilities.

\begin{table}[t]\centering
\captionsetup{width=\linewidth}
\vspace{2mm}
    \caption{Summary of existing python-based STL toolboxes that are publicly available.}
    \label{tab:stltools}
    \begin{tabular}{@{}llll@{}}\toprule
    \textbf{Name} & \textbf{AutoDiff} & \textbf{Vectorize} & \textbf{GPU} \\\toprule
    \stlcgpp\ (Ours) & JAX, PyTorch & \cmark & \cmark\\
    \stlcg \cite{LeungArechigaEtAl2021} & PyTorch & \cmark & \cmark\\
    Argus \cite{Balakrishnan2021}  & \xmark & \xmark & \xmark \\
    stlpy \cite{KurtzLin2022} & \xmark & \xmark & \xmark \\
    PyTeLo \cite{CardonaLeahyEtAl2023} & \xmark & \xmark & \xmark\\
    pySTL \cite{VazquezChanlatte2022} & \xmark & \xmark & \xmark\\
    RTAMT \cite{YamaguchiHoxhaEtAl2023} & \xmark & \xmark & \xmark\\
    \bottomrule
    \end{tabular}
    \vspace{-4mm}
\end{table}

\section{Preliminaries}
\label{sec:prelim}
We provide a brief introduction to STL and related terminologies. See \cite{LeungArechigaEtAl2021,MalerNickovic2004} for a more in-depth description.\vspace{-1mm}
\subsection{Signals and trajectories}
\label{subsec:signals}

\noindent \textbf{Signals and subsignals.} STL formulas are interpreted over one-dimensional \emph{signals} $\signal=(s_0,\ldots,s_T)$, a sequence of scalars sampled at uniform timesteps $\Delta t$ (i.e., continuous-time outputs sampled at finite time intervals) from any system of interest.
Given a signal $\signal=(s_0,\ldots,s_T)$, a \emph{subsignal} is a contiguous fragment of a signal. By default, we assume a subsignal will start at timestep $t$ and end at the last timestep $T$. We denote such a subsignal by $\signal_t =(s_t,\ldots,s_T)$.
If a subsignal ends at a different timestep from $T$, we denote it by $\signal_t^K=(s_t,\ldots,s_K)$. \noindent\emph{Note}: The absence of a sub(super)script on $\signal$ implies that the signal starts (ends) at timestep 0 ($T$).

\noindent \textbf{States, trajectories, and subtrajectories.}
Given a system of interest, let $x_t\in\mathbb{R}^n$ denote the state at timestep $t$. Let $\trajectory=(x_0, \ldots, x_T)$ denote a sequence of states sampled at uniform time steps $\Delta t$.
Similar to how subsignals are defined, we denote a subtrajectory by $\trajectory_t=(x_t, \ldots, x_T)$

\noindent \textbf{From trajectories to signals.}
In this work, we focus specifically on signals computed from a robot's trajectory. As we will see in the next section, core to any STL formula are predicates which are functions mapping state to a scalar value, $\mu : \mathbb{R}^n \rightarrow \mathbb{R}$, with $s_t = \mu(x_t)$, $t = 0, \dots, T$.
A signal can represent, for example, a robot's forward speed.
\vspace{-3mm}
\subsection{Signal Temporal Logic: Syntax and Semantics}
\label{subsec:stl syntax and semantics}
STL formulas are defined recursively according to the following grammar \cite{BartocciDeshmukhEtAl2018,MalerNickovic2004},
\begin{equation}
    \centering
    \begin{tabular}{cc|c|c|c|c}
       $\phi::=$  &  $\top$ &    $\mu_c$   &    $\neg\phi$   &    $\phi \wedge \psi$   &     $\phi\,\Until_{[a,b]}\,\psi$ \\
                  & \small{True} & \small{Predicate} & \small{Not}   & \small{And} & \small{Until}
    \end{tabular}
    \label{eq:STL grammar}
    \vspace{-1mm}
\end{equation}

The grammar \eqref{eq:STL grammar} describes a set of recursive operations that, when combined, can create a more complex formula.
The time interval $[a,b]$ refers to timesteps rather than specific time values. When the time interval is dropped in the temporal operators, it defaults to the entire length of the input signal.
Other commonly used logical connectives and temporal operators can be derived as follows: \emph{Or} ($\phi \vee \psi := \neg(\neg\phi \wedge \neg\psi)$), \emph{Eventually} ($\Eventually_{[a,b]}~\phi := \top~\mathcal{U}_{[a,b]}~\phi$) and \emph{Always} ($\Always_{[a,b]}~\phi := \neg \Eventually_{[a,b]}~\neg\phi$).

A predicate $\mu_c: \mathbb{R}^n \rightarrow \mathbb{R}$ is a function that takes, for example, a robot state and outputs a scalar (e.g., speed).
Then, given a state trajectory $\trajectory=(x_0,\ldots, x_T),\ x_t \in \mathbb{R}^n$, we use the notation \(\trajectory \models \phi\) to denote that the trajectory \(\trajectory\) satisfies \(\phi\) according to the Boolean semantics \eqref{eq:boolean}.

\begin{mybox}{\textbf{Boolean Semantics}}
{\small
\begin{equation}
\begin{aligned}
    \trajectory & \models \mu_c &\Leftrightarrow  \quad& \; \mu(x_0) > c\\
    \trajectory & \models \neg \phi  &\Leftrightarrow  \quad& \; \neg(\trajectory \models \phi)\\
    \trajectory & \models \phi \wedge \psi & \Leftrightarrow  \quad& \; (\trajectory \models \phi) \wedge (\trajectory \models \psi)\\
    \trajectory & \models \Eventually_{[a,b]} \phi & \Leftrightarrow  \quad& \; \exists t \in [a, b] \;\;\mathrm{s.t.} \; \trajectory_t \models \phi\\
    \trajectory & \models \Always_{[a,b]} \phi & \Leftrightarrow  \quad& \; \forall t \in [a, b] \;\;\mathrm{s.t.} \;  \trajectory_t \models \phi\\
    \trajectory & \models \phi \,\Until_{[a,b]}\, \psi & \Leftrightarrow  \quad& \; \exists t \in [a, b] \;\;s.t. \; (\trajectory_t \models \psi)\; \\
    &&& \quad  \wedge \;(\forall \tau \in [0, t], \, \trajectory_\tau \models  \phi)
\end{aligned}
\label{eq:boolean}
\end{equation}
}
\end{mybox}

STL also admits a notion of \emph{robustness}---the \emph{quantitative semantics} in \eqref{eq:quantitative} describes how much a signal satisfies or violates a formula. Positive robustness values indicate satisfaction, while negative robustness values indicate violation.

\begin{mybox}{\textbf{Quantitative Semantics (Robustness Formulas)}}
{\small
\begin{equation}
\begin{aligned}
\rho(\trajectory, \top) & \:= \: \rho_{\max} \qquad \text{where $\rho_{max} > 0$}\\
\rho(\trajectory, \mu_c) & \:= \: \mu(x_0) - c\\
\rho(\trajectory, \neg \phi) & \:= \: -\rho(\trajectory,\phi)\\
\rho(\trajectory, \phi \wedge \psi) & \:= \: \min(\rho(\trajectory,\phi), \rho(\trajectory,\psi))\\
\rho(\trajectory, \Eventually_{[a,b]} \phi) & \:= \: \max_{t \in [a, b]}\rho(\trajectory_t,\phi)\\
\rho(\trajectory, \Always_{[a,b]} \phi) & \:= \: \min_{t \in [a, b]}\rho(\trajectory_t,\phi)\\
\rho(\trajectory, \phi \,\Until_{[a,b]}\psi) & = \max_{t \in [a, b]} \bigg\{\min\left( \min_{\tau\in [0, t]} \rho(\trajectory_{\tau}, \phi), \rho(\trajectory_t, \psi) \right)\bigg\}.
\end{aligned}
\label{eq:quantitative}
\end{equation}
}
\end{mybox}

\subsection{Smooth $\max/\min$ approximation}
It is often desirable to use STL robustness within gradient-based optimization (e.g., training deep neural networks or trajectory optimization).
Since the robustness formulas consist of nested $\max/\min$ operations, smooth approximations $\widetilde{\max}/ \widetilde{\min}$ are often used to aid numerical stability.
Two typical smooth approximations are $\mathrm{softmax}$ and  $\logsumexp$.

{\small
\begin{align*}
    \widetilde{\max}_\mathrm{soft}(\trajectory) &= \sum_{i=1}^N\frac{x_i\exp^{\tau x_i}}{ \sum_{i=j}^N \exp(\tau x_j)},\\
      \widetilde{\max}_\mathrm{LSE}(\trajectory) &= \frac{1}{\tau}\log \sum_{i=1}^N \exp(\tau x_i),\:       \widetilde{\min}(\trajectory) = - \widetilde{\max}(-\trajectory)
\end{align*}
}
The smoothness can be controlled by a temperature parameter $\tau$; as $\tau \rightarrow \infty$, the smooth approximations approach the true $\max/\min$ value. For deep learning, common practice is to anneal the temperature value so that a more accurate smooth $\max/\min$ is used towards the end of the optimization routine.
\noindent Soundness of the smooth approximation may be an important consideration, depending on the specific application. Recent works proposed sound approximations \cite{MehdipourVasileEtAl2019} which \stlcgpp\ can easily support. However, we view these efforts as complementary to this paper.

\subsection{Robustness trace}
As STL formulas are composed of nested operations, evaluating the robustness value necessitates first determining the robustness value for each subtrajectory and subformula.
For instance, consider $\Eventually \phi$ where $\phi$ is the subformula. Evaluating $\rho(\trajectory, \Eventually \phi)$ requires the robustness value of $\phi$ for all subtrajectories, i.e., $\rho(\trajectory_t, \phi)$ for all $t$.
Suppose that $\phi$ is another temporal formula, e.g., $\Always \varphi$, then we would also require $\rho(\trajectory_t, \varphi)$ for all $t$.
Subsequently, we introduce the concept of a \emph{robustness trace}.

\begin{definition}[Robustness trace]
Given a trajectory $\trajectory=(x_0,\ldots,x_T)$ and an STL formula $\phi$,  the \textit{robustness trace} $\robustnesstrace(\trajectory, \phi)$ is a sequence of robustness values of $\phi$ for each subtrajectory of $\trajectory$. That is,
\begin{equation}
\robustnesstrace(\trajectory, \phi) = \rho(\trajectory_0, \phi),\,\rho(\trajectory_1, \phi),\ldots, \rho(\trajectory_T, \phi).
\end{equation}
\end{definition}
\noindent Computing the robustness trace of non-temporal operations is straightforward, as there is no need to loop through time, and for brevity, we omit that description in this paper.

\subsection{Recursive structure of STL formulas}
\begin{figure}
    \vspace{2mm}
    \centering
    \captionsetup{width=\linewidth}
    \includegraphics[width=0.9\linewidth]{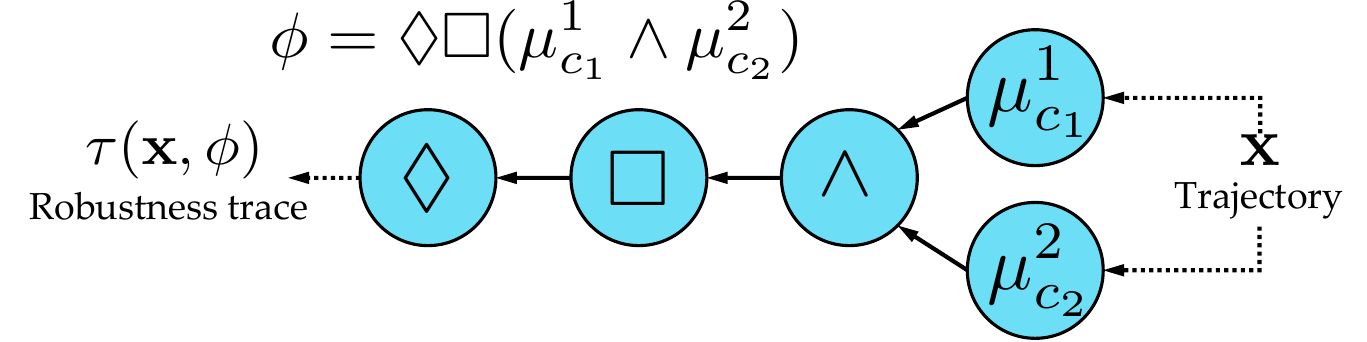}

    \caption{Illustration of the graph structure of an example STL formula. Input trajectories are first passed through the predicates and then each subformula according to the formula structure.}
    \label{fig:example graph}
    \vspace{-6mm}
\end{figure}

 To compute the robustness value of any STL formula $\phi$ with arbitrary formula depth\footnote{Assuming the formula consists of at least one temporal operation. Although STL formulas without temporal operators are possible, it is the temporal operations that make computing robustness challenging.}, we must first calculate the robustness trace of its subformula(s), and so forth, since STL formulas are constructed recursively.
 Fig.~\ref{fig:example graph} illustrates how the trajectories are passed through each operation of an STL formula to compute the robustness trace of each subformula.
As such, we need to develop the ``building blocks'' for computing robustness corresponding to each STL operation. Then, by stacking these operations according to the formula structure, we can compute the robustness trace for \textit{any} STL formula regardless of formula depth.

\subsection{\stlcg: Recurrent computation}
We briefly outline the recurrent operations underlying \stlcg; for more details, we refer the reader to \cite{LeungArechigaEtAl2021}.
Illustrated in Fig.~\ref{fig:hero} (right), \stlcg\ utilizes the concept of dynamic programming to calculate the robustness trace. The input signal is processed \textit{backward} in time, and a \textit{hidden state} is maintained to store the information necessary for each recurrent operation at each time step. The choice of recurrent operation depends on the temporal STL formula (either a $\max$ or $\min$).
The size of the hidden state depends on the time interval of the temporal operator and is, at most, the length of the signal.
Although the use of a hidden state to summarize past information helps reduce space complexity, the recurrent operation leads to slow evaluation and backpropagation due to sequential dependencies.
Next, we present a masking-based approach that bypasses the sequential dependency, leading to faster computation times.

\section{Masking approach for temporal operations}
We propose \stlcgpp, a \textit{masking} approach to compute STL robustness traces. Fig.~\ref{fig:hero} illustrates the masking approach, highlighting the idea that each value of a robustness trace is computed \textit{simultaneously}, rather than sequentially, as we saw with \stlcg.
Mirroring the concept of attention masks in transformer architectures \cite{VaswaniShazeerEtAl2017}, we introduce a mask $\masked$ to select relevant parts of a signal that can be later processed simultaneously rather than sequentially.
In the rest of this section, we describe the masking operation underpinning \stlcgpp\ and several properties.

\subsection{Notation and masking operation}
Assume we have a 1D input signal $\signal$ of length $T+1$, and let $K$ denote the number of time steps contained (inclusive) in the time interval $[a,b]$ given an STL temporal operator.
A mask $\masked$ is a 2 or 3 dimensional array, depending on the temporal operator, whose entries are either $1$ or $0$.
A mask $\masked$ is applied via element-wise multiplication to an array $\unrolled$ of the same size. E.g., for a 2D mask, if $\masked_{ij}=0$, the corresponding entries $\unrolled_{ij}$ will be ignored, or is \textit{masked out}. We also replace the entries masked out with an arbitrary value, typically a large positive or negative value.
Let $\masked \odot_{\pm} \unrolled$ denote the application of mask $\masked$ on $\unrolled$ and masked out by either a large positive ($\odot_{+}$) or negative ($\odot_{-}$) number.

\subsection{Eventually and Always operations}
\label{subsec:eventually always}
\label{subsec: ev alw mask procedure}
For brevity, we describe the procedure for the Eventually operation but note that the procedure for the Always operation is almost identical except a $\min$ operation is used instead of $\max$.
Consider the formula $\psi = \Eventually_{[a,b]} \phi$, and let $\signal=\robustnesstrace(\trajectory, \phi)$ denote the robustness trace of $\phi$. Specifically, $s_t = \rho(\trajectory_t, \phi)$. Here, $\phi$ is an STL subformula, typically representing a predicate or a more complex temporal logic expression.
Then, we seek to find the robustness trace,
\vspace{-1mm}
\[\robustnesstrace(\trajectory, \Eventually_{[a,b]} \phi) = \biggl\{ \max_{i\in[a,b]} s_{t+i}  \biggl\}_{t=0}^{T}\]

Intuitively, we want to slide a time window $[a,b]$ along $\signal$ and take the $\max$ of the values within the window.
Rather than performing this set of operations sequentially, we \textbf{first} ``unroll'' the sequential operation---turning the 1D signal into a 2D array, \textbf{second}, we apply a 2D mask to mask out the irrelevant entries dictated by the time window, and \textbf{third}, we use the $\max$ operation over the unmasked values.

\noindent\textbf{Step 1:``Unrolling'' the signal.}
First, we \textit{pad} the end of the signal by the size of the time window $K$; the reason will be apparent in Step 3. The padding value $\bar{s}$ can be set by the user, such as extending the last value of the signal, or setting it to a large negative number.
Then, we turn the padded signal into a 2D array $\unrolled$ by repeating the signal along the second dimension $T$ times, resulting in a $(T+K) \times T$ 2D array. An example is provided in \eqref{eq:masked 2D array}.

\noindent\textbf{Step 2: Construct 2D mask.}
We construct a mask $\masked$ that will be applied to the unrolled signal $\unrolled$.
The mask $\masked$ comprises of two sub-masks: (i) subsignal mask $\masked_\mathrm{subsig}$ and (ii) time interval mask $\masked_\mathrm{time}$.

\noindent\textit{Subsignal mask} $\masked_\mathrm{subsig}$: This mask incrementally masks out the start of the signal as we move along the horizontal (second) dimension of $\unrolled$. Essentially, $\masked_\mathrm{subsig}$ masks out the upper triangular region with an offset of one to exclude the diagonal entries. After applying the $\masked_\mathrm{subsig}$ on $\unrolled$, the columns of the unmasked entries correspond to all the subsignals of $\signal$. The entries masked out by $\masked_\mathrm{subsig}$ are denoted by the red shaded entries in \eqref{eq:masked 2D array}.

\noindent\textit{Time interval mask} $\masked_\mathrm{time}$: This mask masks out all the entries outside of time interval $[a,b]$ for each subsignal. $\masked_\mathrm{time}$ consists of an off-diagonal strip that masks out entries \textit{before} the start of the interval (determined by $a$), and a lower triangular matrix with an offset that masks out entries \textit{after} the interval (determined by $b$).
The entries masked out by $\masked_\mathrm{time}$ are denoted by the blue shaded entries in \eqref{eq:masked 2D array}.

\noindent\textit{Final mask} $\masked$: The final mask is $\masked = \masked_\mathrm{subsig} + \masked_\mathrm{time}$, resulting in a mask that retains all the entries within the time interval $[a,b]$ for each sub-signal (unshaded entries in \eqref{eq:masked 2D array}).

\noindent\textbf{Step 3: Apply 2D mask and $\max$ operation.}
Given $\unrolled$ and $\masked$ from the previous steps, we can compute $\masked \odot_{-}\unrolled$, similar to what is done in the mask attention mechanism in transformer architectures. Then, we apply the $\max$ operation along each column.
Note the padding we did back in step 1 becomes relevant when the time interval exceeds the length of the subsignal. This is similar to the approach taken in \cite{LeungArechigaEtAl2021} to handle incomplete signals.
For the Always operation, $\min$ and $\masked\odot_{+}\unrolled$ are used instead.

\begin{example}
Consider the STL formula $\psi = \Eventually_{[1,3]} (s > 0)$ and a signal with 8 time steps, $\signal=[s_0,\ldots, s_7]$.
The corresponding unrolled 2D array is shown \eqref{eq:masked 2D array} with padding value $\bar{s}$, and the red and blue shaded entries denote the entries masked out by $\masked_\mathrm{subsig}$ and $\masked_\mathrm{time}$, respectively.

\begin{equation}
{\small
\begin{bNiceMatrix}[margin]
\Block[fill=blue!15,rounded-corners]{1-1}{} s_0 & \Block[fill=red!15,rounded-corners]{1-7}{}
      s_0 & s_0 & s_0 & s_0 & s_0 & s_0 & s_0 \\
s_1 & \Block[fill=blue!15,rounded-corners]{1-1}{}s_1 & \Block[fill=red!15,rounded-corners]{1-6}{}
            s_1 & s_1 & s_1 & s_1 & s_1 & s_1 \\
s_2 & s_2 & \Block[fill=blue!15,rounded-corners]{1-1}{}s_2 & \Block[fill=red!15,rounded-corners]{1-5}{}
                  s_2 & s_2 & s_2 & s_2 & s_2 \\
s_3 & s_3 & s_3 & \Block[fill=blue!15,rounded-corners]{1-1}{}s_3 & \Block[fill=red!15,rounded-corners]{1-4}{}
                        s_3 & s_3 & s_3 & s_3 \\
\Block[fill=blue!15,rounded-corners]{1-1}{}s_4 & s_4 & s_4 & s_4 & \Block[fill=blue!15,rounded-corners]{1-1}{}s_4 & \Block[fill=red!15,rounded-corners]{1-3}{}
                              s_4 & s_4 & s_4 \\
\Block[fill=blue!15,rounded-corners]{1-2}{}s_5 & s_5 & s_5 & s_5 & s_5 & \Block[fill=blue!15,rounded-corners]{1-1}{}s_5 & \Block[fill=red!15,rounded-corners]{1-2}{}
                                    s_5 & s_5 \\
\Block[fill=blue!15,rounded-corners]{1-3}{}s_6 & s_6 & s_6 & s_6 & s_6 & s_6 & \Block[fill=blue!15,rounded-corners]{1-1}{}s_6 & \Block[fill=red!15,rounded-corners]{1-1}{}
                                          s_6 \\
\Block[fill=blue!15,rounded-corners]{1-4}{} s_7 & s_7 & s_7 & s_7 & s_7 & s_7 & s_7 & \Block[fill=blue!15,rounded-corners]{1-1}{}s_7 \\
\Block[fill=blue!15,rounded-corners]{1-5}{} \bar{s} & \bar{s} & \bar{s} & \bar{s} & \bar{s} & \bar{s} & \bar{s} & \bar{s} \\
\Block[fill=blue!15,rounded-corners]{1-6}{} \bar{s} & \bar{s} & \bar{s} & \bar{s} & \bar{s} & \bar{s} & \bar{s} & \bar{s} \\
\Block[fill=blue!15,rounded-corners]{1-7}{} \bar{s} & \bar{s} & \bar{s} & \bar{s} & \bar{s} & \bar{s} & \bar{s} & \bar{s} \\
\end{bNiceMatrix}
}
\label{eq:masked 2D array}
\end{equation}

After filling the masked entries with a large negative value, and taking the $\max$ column-wise.
Let $\signal = [0, 1, 2, 3, 4, 5, 6, 7]$ and $\bar{s} = 7$, then $\tau(\signal, \psi)=[3, 4, 5, 6, 7, 7, 7, 7]$.
If we set $\bar{s} = -10^5$, then $\tau(\signal, \psi)=[3, 4, 5, 6, 7, -10^5, -10^5, -10^5]$.

\label{eg:masking}
\end{example}

\subsection{Until operation}
\label{subsec:until}

We apply a similar masking strategy for the Until operation, following a similar three-step process described in Sec.~\ref{subsec: ev alw mask procedure}. The main differences are (i) the signal needs to be ``unrolled'' into three dimensions, and (ii) the operations in Step 3 correspond to the Until robustness formula.

Consider the formula $\phi\, \Until_{[a,b]} \psi$ where $\phi$ and $\psi$ are STL subformulas.
Let $\signal^\phi=\robustnesstrace(\trajectory, \phi)$ and $\signal^\psi=\robustnesstrace(\trajectory, \psi)$ denote the robustness trace of $\phi$ and $\psi$ respectively for trajectory $\trajectory$.
Specifically, $s_t^\phi= \rho(\trajectory_t, \phi)$ and $s_t^\psi=\rho(\trajectory_t, \psi)$.
Then, we seek to find the robustness trace,

\vspace{-5mm}
{\small
\begin{equation*}
\robustnesstrace(\trajectory, \phi\,\Until_{[a,b]}\psi) = \Bigg\{\max_{i \in [a, b]} \bigg\{\min\left( \min_{\tau\in [0, i]} s_{t+\tau}^\phi,\ s_{t+i}^\psi) \right)\bigg\}\Bigg\}_{t=0}^{T}
\end{equation*}
}

\noindent\textbf{Step 1:``Unrolling'' the signal.}
Given a signal $\signal$, we construct a 3D array by repeating the signal along the first and second dimensions, resulting in a $(T+K)\times T \times K$ array. This unrolling operation on $\signal^\phi$ and $\signal^\psi$ results in $\unrolled^\phi$ and $\unrolled^\psi$.

\noindent\textbf{Step 2: Construct 3D mask.}
In the Until robustness formula, we need $\min_{\tau\in [0, i]} \signal_{t+\tau}^\phi$ and $\signal_{t+i}^\psi$ for $i\in [a,b]$ and for all $t$.
Given $i$, notice that $\rho(\trajectory_t, \Always_{[0,i]} \phi) = \min_{\tau\in [0, i]} \signal_{t+\tau}^\phi$ and $\rho(\trajectory_t, \Always_{[i,i]} \psi) = \signal_{t+i}^\psi$ which we can compute given the procedure outlined in Sec.~\ref{subsec: ev alw mask procedure}.
However, the outer $\max$ requires us to consider all $i\in[a,b]$. Thus for each formula $\Always_{[0,i]} \phi$ and $\Always_{[i,i]} \psi$, we stack all the 2D masks for each value of $i\in [a,b]$ across the third dimension, resulting in two 3D masks, $\masked^\phi$ and $\masked^\psi$ of size $(T+K)\times T \times K$.

\noindent\textbf{Step 3: Apply the Until robustness operations.}
We can now apply all the operations needed to compute the robustness trace for the Until operation.
After Step 2, we have $\unrolled^\phi$, $\unrolled^\psi$, and the 3D masks $\masked^\phi$ and $\masked^\psi$. We can compute $\masked^\phi \odot_{+} \unrolled^\phi$ and $\masked^\psi \odot_{+}\unrolled^\psi$. Then, we can apply the following set of operations (where the first dimension $=1$).

\vspace{-5mm}
{\small
\begin{align*}
    S_1 &= \overbrace{\underset{\mathrm{dim}=1}{\mathrm{stack}}\left( \left[\min_{\mathrm{dim}=1} \masked^\phi \odot_{+} \unrolled^\phi,  \min_{\mathrm{dim}=1} \masked^\psi \odot_{+}\unrolled^\psi \right] \right)}^{\text{Size } [2 \times T \times K]}\\
    S_2 &= \min_{\mathrm{dim}=1} S_1, \qquad (S_2 \text{ has size } [T \times K])\\
    S_3 &= \max_{\mathrm{dim}=2} S_2, \qquad (S_3 \text{ has size } [T])
\end{align*}
}

We see that this set of operations is relatively simple, a series of $\min$ and $\max$ operations along various dimensions, and these operations remain the same regardless of the choice of time intervals, unlike the recurrent approach.

\subsection{Dependence on smooth approximations}

Both masking and recurrent approaches output the correct robustness values when the true $\min/\max$ operations and $\logsumexp$ approximation are used. However, if the $\softmax/\softmin$ approximation is used with a recurrent approach, the output robustness and gradient values do not correctly reflect the correct values.
This is because, when recurrently applying $\softmin/\softmax$, values applied earlier in the recurrence will be ``softened'' more than values applied later in the recurrence. Not only does nesting the $\softmin/\softmax$ operations give a very poor approximation of the true $\min/\max$ value, but it also artificially reduces the gradient for values passed earlier in the recurrence.
With the $\logsumexp$ approximation, applying it recursively over multiple values is the same as applying it once over all the values. That is, $\widetilde{\max}_\mathrm{LSE}([\widetilde{\max}_\mathrm{LSE}(\trajectory), y]) = \widetilde{\max}_\mathrm{LSE}([\trajectory, y])$ where as $\widetilde{\max}_\mathrm{soft}([\widetilde{\max}_\mathrm{soft}(\trajectory), y]) \neq \widetilde{\max}_\mathrm{soft}([\trajectory, y])$. This issue is not observed when using the masking-based approach due to the way subsignals are processed without recurrence.

\begin{table*}[t]\centering
    \vspace{2mm}
    \caption{Summary of STL specifications used for evaluation inspired by existing literature. $I$ denotes the time interval over which the temporal operators apply.}
    \label{tab:stl_formulas}
    \begin{tabular}{@{}llll@{}}\toprule
    \textbf{Spec} & \textbf{Depth} &\textbf{Description} & \textbf{STL Formula}\\\toprule
    $\phi_1$ & 0 &
    Invariance &
    $\Always (\varphi \wedge \psi)$ \\
    $\phi_2$ & 1 &
    Stabilization&
    $\Eventually \Always (\varphi \wedge \psi)$ \\
    $\phi_3$ & 1 &
    Strict ordering &
    $\varphi\ \Until\ \psi$ \\
    $\phi_4$ & 3 &
    Sequenced visit pattern &
    $\Eventually_{I}(\varphi_4 \land \psi_4 \land
                \Eventually_{I}(\varphi_3 \land \psi_3 \land
                \Eventually_{I}(\varphi_2 \land \psi_2 \land
                \Eventually_{I}(\varphi \land \psi))))$ \\

    $\phi_5$ & 2 &
    Sequenced visit + stabilization &
    $\Eventually_{I}((\varphi_2 \land \psi_2) \land
                \Eventually_{I}(\Always_{I}(\varphi \land \psi)))$ \\

    $\phi_6$ & 0 &
    Reach regions in any order &
    $\Eventually_{I}(\varphi_9 \land \psi_9) \land
                \ldots \land
                \Eventually_{I}(\varphi_0 \land \psi_0)$ \\
    \bottomrule
    \end{tabular}
    \vspace{-4mm}
\end{table*}

\subsection{Practical benefits}
\stlcgpp\ offers several practical benefits over \stlcg. We highlight several reasons why this is the case.
\textbf{(i) Software design.} The recurrent approach required signals to be passed \textit{backward} in time. This increased the complexity of software design and was not often intuitive for end users. With the masking approach, there is no need to reverse the signal.
\textbf{(ii) Static graph structure.} With the recurrent approach, the choice of time interval would fundamentally change the dimension of hidden states, whereas the graph structure remains the same for the masking approach.
Keeping the same graph structure, even if the intervals change, is essential in compilation.
\textbf{(iii) Vectorization and just-in-time compilation.} Related to the second point above, the static graph structure afforded by the masking approach enables the ability to easily \textit{vectorize} and \textit{just-in-time compile} (JIT) the computation over not just various signal inputs but also various time intervals. The ability to vectorize and JIT over different time intervals can be particularly useful in formula mining applications where we may want to evaluate robustness formulas with different time intervals simultaneously.
\textbf{(iv) Conformity to popular deep learning libraries.} By leveraging AD libraries that empower popular frameworks, \stlcgpp\ becomes accessible to the broader community and seamlessly integrates into existing tools.

\section{Computational properties of \stlcgpp}
In this section, we analyze the computational properties of the approaches \stlcgpp \ (masking-based) and \stlcg \ (recurrent-based) by measuring the computation time required to compute robustness values and their gradient.
We seek to answer the following research questions.

\noindent \textbf{RQ1}: Does \stlcgpp\ compute robustness traces faster than \stlcg\ as measured by median computation time?

\noindent \textbf{RQ2}: How does \stlcgpp's computation time scale with sequence length compared to \stlcg?

\begin{figure*}[t]
    \centering
    \captionsetup{width=\linewidth}
    \includegraphics[width=\linewidth]{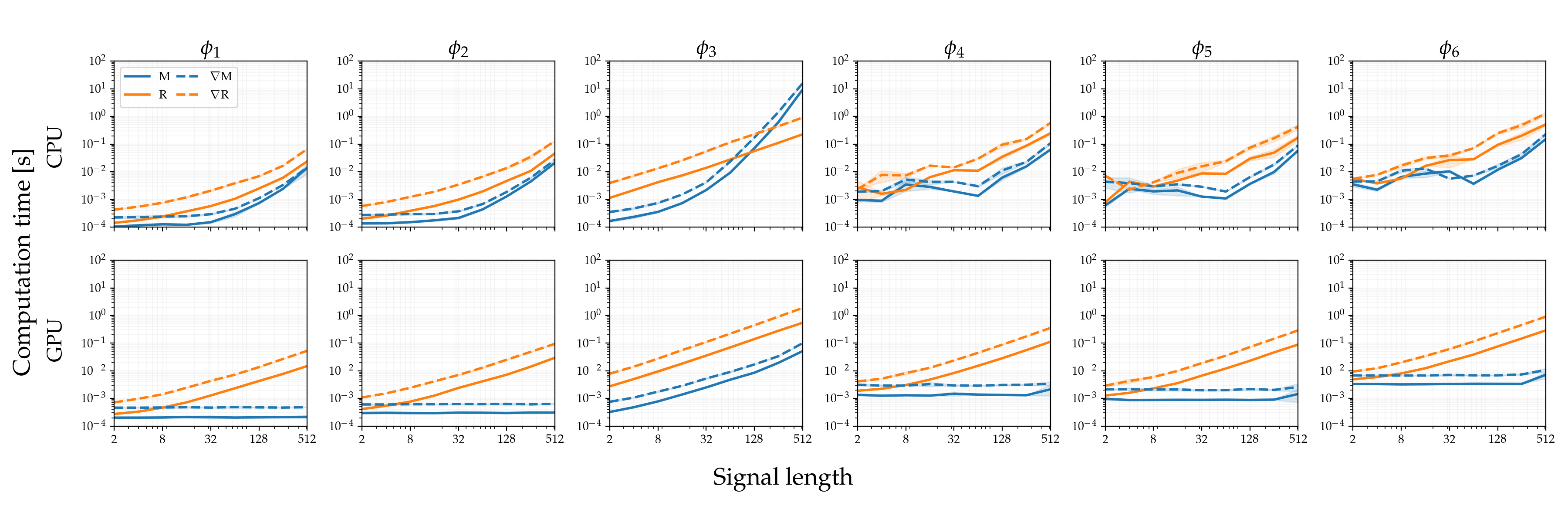}
    \vspace{-8mm}
    \caption{
    Comparison of computation time using PyTorch for masking (M, blue) and recurrent (R, orange) approach on CPU (top row) and GPU (bottom row) as signal length increases, across six STL formulas. Solid lines denote robustness computation; dashed lines denote gradient evaluation. Results using JAX can be found on the website.
    }
    \label{fig:computation times}
    \vspace{-2mm}
\end{figure*}

We perform experiments on CPU and GPU. Since \stlcgpp\ (masking) involves large matrix computations, we anticipate \stlcgpp\ to scale favorably on a GPU. As \stlcg\ (recurrent) utilizes a recurrent structure, it scales with sequence length.
Tab.~\ref{tab:stl_formulas} describes six different STL formulas with varying complexities \cite{HeBartocciEtAl2022} that we test on.
We evaluate computation time for increasing signal lengths up to $T=512$ time steps with a batch size of $8$.
We present our results in Fig.~\ref{fig:computation times} and Tab.~\ref{tab:computational improvement} and make the following observations.

\noindent \textbf{CPU backend.}
We observe that \stlcgpp\ generally achieves lower computation times than \stlcg. The exception is in $\phi_3$, where \stlcgpp\ is slower than \stlcg\ for longer sequences. This is because the space complexity for the Until operation is $\mathcal{O}(T^3)$.
In Tab.~\ref{tab:computational improvement}, we see that \stlcgpp\ on JAX struggled with $\phi_3$ for long sequence lengths, but was fine with PyTorch. We hypothesize that it is due to how JAX allocates memory, especially during just-in-time compilations. Additionally, both libraries see increasing computation time with sequence length. For \stlcg\, the computation time increase is expected since the time complexity is $\mathcal{O}(T)$. For \stlcgpp, the increased computation time can be explained by the space complexity $\mathcal{O}(T^2)$ (except for $\phi_3$), and such computations are not handled as efficiently on a CPU.

\noindent \textbf{GPU backend.}
\stlcgpp\ outperforms \stlcg\ for all formulas.  Additionally, \stlcgpp\ exhibits essentially constant computation time except for $\phi_3$, which has $\mathcal{O}(T^3)$ space complexity. From Tab.~\ref{tab:computational improvement}, we see that \textbf{\stlcgpp\ on GPU provides around 95\% and 85\% improvement on JAX and PyTorch, respectively}. For \stlcg, moving from CPU to GPU gives virtually the same scaling, with no significant improvement/reduction in computation times.

\begin{table}[t]\centering
    \vspace{-3mm}
    \captionsetup{width=\linewidth}
    \caption{Relative computation time of masking approach compared to recurrent approach. Median value across different signal lengths. Lower is better.}
    \label{tab:computational improvement}
    \begin{tabular}{@{}c|cc|cc@{}}\toprule
    \textbf{Device} & \textbf{CPU} & \textbf{GPU} & \textbf{CPU} & \textbf{GPU} \\ \midrule
    \textbf{Formula} & \multicolumn{2}{c|}{JAX + JIT} & \multicolumn{2}{c}{PyTorch} \\ \toprule
    $\phi_1$  &             $ -37.99\% $ & $ -93.43\% $ & $ -76.05\% $ & $ -85.76\% $ \\
    $\phi_2$  & $ -51.51\% $ & $ -96.65 $\% & $ -82.50\% $ & $ -89.03\% $ \\
    $\phi_3$  & $ 822.01\% $ & $ -91.31 \% $ & $ -88.22 \% $ & $ -94.15\% $ \\
    $\phi_4$ & $-68.22\%$ & $-96.00\%$ & $-73.48\%$ & $-77.91\%$ \\
    $\phi_5$ & $-68.58\%$ & $-95.32\%$ & $-67.91\%$ & $-80.86\%$ \\
    $\phi_6$ & $-72.93\%$ & $-98.11\%$ & $-54.14\%$ & $-84.36\%$ \\
    \bottomrule
    \end{tabular}
    \vspace{-6mm}
\end{table}

\section{Smoothing time intervals}
\label{sec: differentiable time}

By using the masking operation to capture parts of the signal within the specified time interval, we can build a smooth approximation of the mask and \textit{differentiate} robustness values with respect to the parameters of the mask that determine which values are selected.
As such, it becomes possible to perform gradient descent over time interval parameters to find, for example, a time interval that best explains time-series data. We showcase two examples in Sec.~\ref{sec:examples} utilizing this new capability.
Notably, this differentiation was not possible with \stlcg\ since the choice of time intervals impacted the size of the hidden state used within the recurrent computations.

Rather than using a mask with $0$ and $1$'s as described in Sec.~\ref{subsec:eventually always} and \ref{subsec:until}, we use the following smooth mask approximation.
For a sequence length of $T$, we have,

\vspace{-5mm}
{\small
\begin{equation}
    \smoothmask(i; a, b, c, \epsilon) = \max(\sigma(c(i - aT))  - \sigma(c(i - bT)) - \epsilon, 0),
    \label{eq:smooth mask}
\end{equation}
}

\begin{figure}[t]
    \centering
    \vspace{-3mm}
    \captionsetup{width=\linewidth}
    \includegraphics[width=\linewidth]{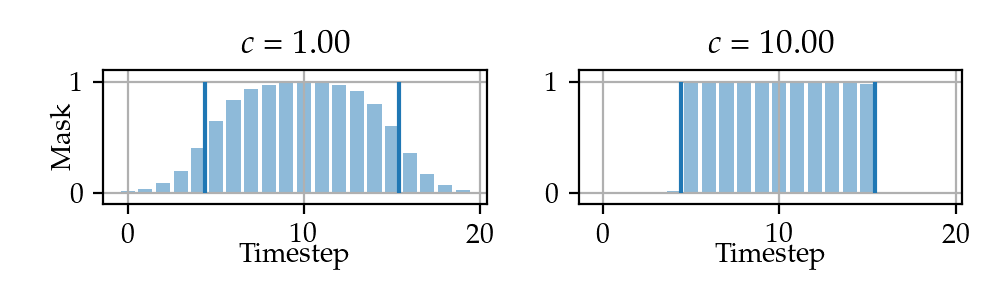}
    \vspace{-8mm}
    \caption{Smooth mask for various values of smoothing parameter $c$.}
    \label{fig:smooth mask}
    \vspace{-7mm}
\end{figure}

\vspace{-1mm}
\noindent where $i$ is the time index, $\sigma(x)$ is the sigmoid function, the parameters $0 \leq a < b \leq 1$ denote the fraction along the signal of the start and end of the time interval, $c$ denotes the mask smoothing parameter, and $\epsilon$ denotes a user-defined tolerance.
Fig.~\ref{fig:smooth mask} shows a visualization of the smooth mask and how the smoothing parameter $c$ affects the mask.

When optimizing over time interval parameters, we can anneal the smooth mask parameter $c$ to help with convergence to a (local) optima on a generally nonlinear loss landscape (see Sec.~\ref{subsec:pstl time interval} for more details).
As noted before, a benefit of the masking approach is that the operations can be easily vectorized. We can simultaneously evaluate the robustness for multiple time intervals and perform gradient descent on multiple values.
As such, we can envision a use case where we perform a coarse global search via sampling and then a local refinement via gradient descent.

For reference, using the vectorized mapping function in JAX+JIT, the computation time to evaluate $90,000$ different values for $(a,b)$ with a signal of length $T=20$ for the formula $\Always_{[aT, bT]} (s > 0)$ is
$1.87 \pm 0.0164$ms on a M2 MacBookPro. This averages to around $20.78\mu$s per time interval value.
In contrast, (sequentially) searching over all possible valid time intervals for a signal length of 20 (i.e., 190 intervals with integer interval limits) using the recurrent approach takes about
$5.25  \pm 0.0497$s,  or roughly $27.6$ms each evaluation.
This means that \textbf{\stlcgpp\ offers more than 1000$\boldsymbol{\times}$ improvement in computation time} when evaluating STL robustness over multiple time interval values.

\section{Robotics-related applications}
\label{sec:examples}
We showcase a variety of use cases that demonstrate the computational advantages of the masking approach. We developed \textit{two} Python \stlcgpp\ libraries, one in JAX \cite{JAX2018} and another in PyTorch \cite{PaszkeGrossEtAl2017} and discuss how \stlcgpp\ opens up new possibilities for robot planning and control.

\subsection{Trajectory planning with suboptimal STL specifications.}
\label{subsec:traj opt}
\begin{figure}
     \centering
    \captionsetup{width=\linewidth}
     \begin{subfigure}[t]{0.23\textwidth}
         \centering
         \includegraphics[width=\linewidth]{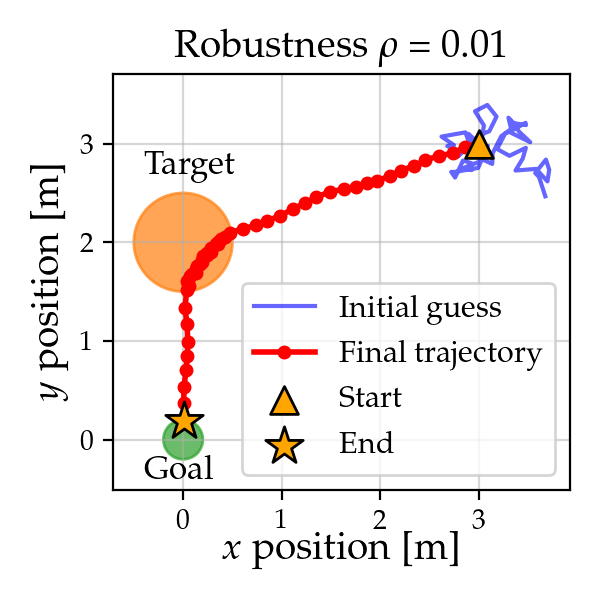}
         \vspace{-5mm}
         \caption{Trajectory with optimized time interval in orange target region.}
         \label{fig:traj opt diff time}
     \end{subfigure}
     \begin{subfigure}[t]{0.23\textwidth}
         \centering
         \includegraphics[width=\linewidth]{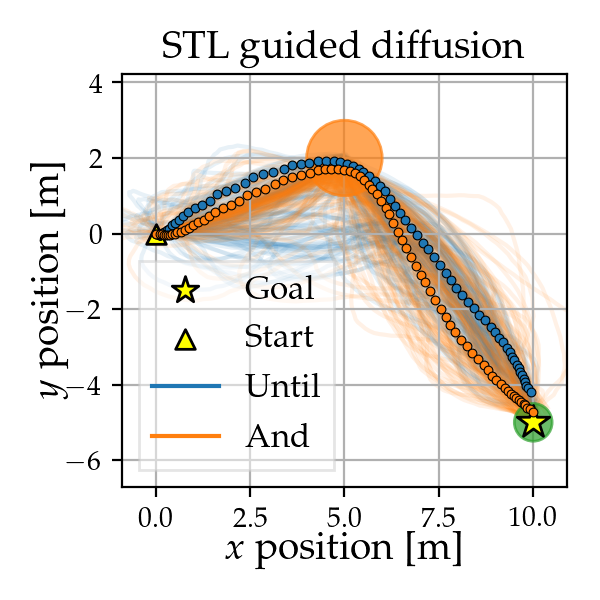}
         \vspace{-5mm}
        \caption{Generated trajectories using STL guided DDIM.}
        \label{fig:stl ddim}
    \end{subfigure}
    \caption{Trajectories generated using \stlcgpp\ with STL robustness in the planning objective.}
    \vspace{-4mm}
\end{figure}

Trajectory optimization is essential in robotic systems, enabling precise navigation and task execution under complex constraints. These trajectories must balance user-defined goals with feasibility and efficiency. Since goal misspecification can lead to erroneous behavior, infeasibility, or subpar performance, we show how suboptimal goals (captured through STL specifications) can be refined in addition to solving for an optimal trajectory.
Consider a trajectory optimization problem where we would like to reach a goal region \rounded{$^.$}{ForestGreen} while visiting a target region \rounded{$^.$}{YellowOrange} within a fixed time horizon. Additionally, the visit duration of \rounded{$^.$}{YellowOrange} is to be maximized.
Consider a specification $\phi = \Always_{[aT, bT]} (\text{inside }\rounded{$^.$}{YellowOrange}) \wedge \Eventually (\text{inside }\rounded{$^.$}{ForestGreen})$ which specifies a time window that the trajectory should be inside \rounded{$^.$}{YellowOrange}.
We cast this trajectory planning problem as an unconstrained optimization problem,
\vspace{-3mm}

{\small
\begin{align*}
&\min_{\controls, a, b} \gamma_1 J_\mathrm{STL}(\trajectory) + \gamma_2 J_\mathrm{I}(a,b) + \gamma_3 J_\mathrm{lim}(\controls)  + \gamma_4 J_\mathrm{eff}(\controls),\\
&J_\mathrm{STL}(\trajectory) = \mathrm{ReLU}(-\rho(\trajectory, \phi)), \quad J_\mathrm{I}(a,b) = \exp(2(\tilde{I} - b + a)),\\
&J_\mathrm{lim}(\controls) = \frac{1}{T} \sum_{t=0}^{T-1}\mathrm{ReLU}(\| \controls_t \|_2 - \bar{u}), \: J_\mathrm{eff}(\controls) = \frac{1}{T} \sum_{t=0}^{T-1}\| \controls_t\|_2^2,
\end{align*}
}

\noindent where $\trajectory$ is the state trajectory from executing $\controls$ with single integrator discrete-time dynamics.
We use a timestep of $\Delta t= 0.1$, $T=51$, $\bar{u}=2$ is the system's maximum control limits, and $\tilde{I}=0.2$ is a nominal (normalized) time interval size that we would like to improve upon.
We randomly initialized the control inputs and $(a,b)_\mathrm{init} = (0.14, 0.82)$.
Using coefficients $\gamma_1=1.1, \gamma_2=0.05, \gamma_3=2, \gamma_4=0.5$, the resulting solution is shown in Fig.~\ref{fig:traj opt diff time} with final values $(a,b)_\mathrm{final} = (0.37, 0.84)$. The final optimized STL formula is $\phi = \Always_{[20, 43]} (\text{inside }\rounded{$^.$}{YellowOrange}) \wedge \Eventually (\text{inside }\rounded{$^.$}{ForestGreen})$. Using JAX, each gradient step took about
$92.3 \pm 3.99\mu$s on an M2 MacBookPro.
With unconstrained multiobjective optimization, selecting objective weights that achieve desirable behavior can be tedious. With \stlcgpp, we can perform gradient descent computation over various coefficient values simultaneously and then select the best one.

\subsection{Deep generative modeling: STL-guided diffusion policies }

We demonstrate the use of STL specifications for guided diffusion models \cite{SongMengEtAl2021}, a recent type of deep generative models that present a promising approach for robot policy learning and behavior generation \cite{JannerDuEtAl2022}.
We build upon \cite{MizutaLeung2024}, which uses Control Barrier and Lyapunov functions (CBFs and CLFs) to guide the denoising process of a diffusion model for safe control sequence generation. In this application, we use STL robustness in place of CBFs and CLFs as the guidance function. We trained a Denoising Diffusion Implicit Model (DDIM) model \cite{SongMengEtAl2021}, generating trajectories with 80 time steps.

We used 200 denoising steps and a batch size of 128. We generated trajectory samples using two different STL guidance functions,
\vspace{-1mm}
\begin{align*}
    \phi_\Until &= \Eventually \Always_{[0,10]} (\text{inside } \rounded{$^.$}{YellowOrange}) \ \Until\ \Eventually \Always_{[0,5]}(\text{inside } \rounded{$^.$}{ForestGreen}) \\
    \phi_\wedge &= \Eventually \Always_{[0,10]} (\text{inside } \rounded{$^.$}{YellowOrange}) \wedge \Eventually_{[60,80]} \Always_{[0,5]} (\text{inside } \rounded{$^.$}{ForestGreen}).
\end{align*}

The resulting trajectories are shown in Fig.~\ref{fig:stl ddim}. Although guided diffusion does not guarantee that the STL specifications are strictly satisfied, we observe a satisfaction rate of $52.34\%$, $62.50\%$ for $\phi_\Until$ and $\phi_\wedge$, respectively. This shows how \stlcgpp\ can enhance diffusion policies to promote safer behavior generation.

\subsection{Machine learning: STL parameter mining from data}
\label{subsec:pstl time interval}

\begin{figure*}
    \centering
    \captionsetup{width=\linewidth}
    \includegraphics[width=\textwidth]{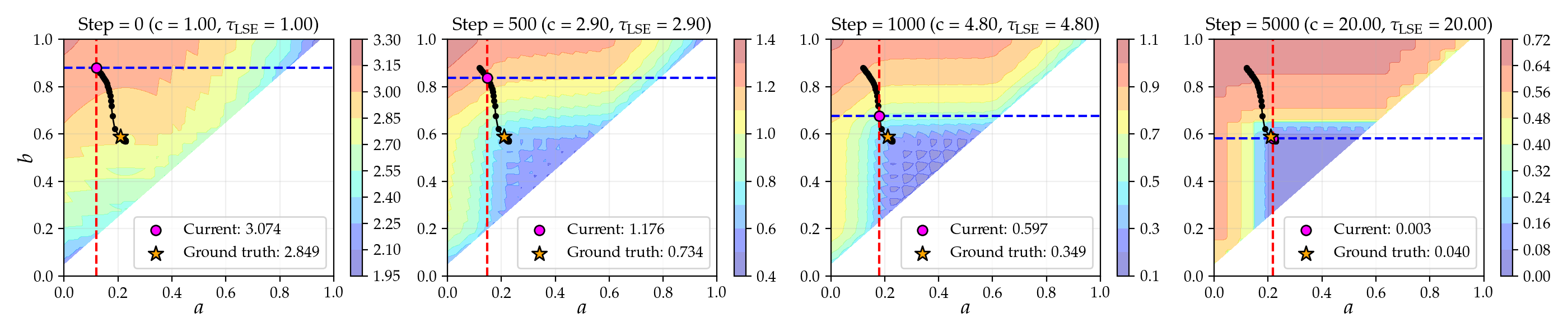}
    \vspace{-7mm}
    \caption{The masking approach enables a smooth approximation of time intervals, enabling the use of gradient descent over time interval parameters.
    The contour plots show the loss landscape given by \eqref{eq:pSTL problem} as a function of $a,b$, the normalized time interval limits. The solutions at gradient descent steps 0, 500, 1000, and 5000 are shown. Note that the smooth time mask parameter and approximate $\max / \min$ temperature values are annealed over the steps.}
    \label{fig:time interval gradient descent}
    \vspace{-5mm}
\end{figure*}

Formal STL specifications play a crucial role in building robust robot systems, contributing to controller synthesis \cite{KapoorKangEtAl2024}, fault localization \cite{BartocciManjunathEtAl2021}, and anomaly detection and resolution \cite{KangGanlathEtAl2024}. However, specifications are not often readily available, making the task of mining specifications from data a vital problem to study \cite{HeBartocciEtAl2022}.
In this example, we perform STL specification mining from data. Specifically, we leverage the differentiability and vectorization over smooth time intervals to find a time interval that best fits the observed data.

Consider a dataset of (noisy) signals with $T=20$ timesteps. The signals have value $1$ between the normalized time interval of $(a,b)=(0.23, 0.59)$ and zero elsewhere. Some noise is added around $(a,b)$ and to the signal itself.
The goal is to learn the largest time interval $[\bar{a}, \bar{b}]$ such that the signals from the dataset satisfy $\phi = \Always_{[\bar{a}, \bar{b}]} (s > 0)$. We frame this STL mining problem as an optimization problem,

\vspace{-4mm}
{\small
\begin{equation}
    \min_{0 \leq a < b \leq 1} \: \frac{1}{N} \sum_{\signal\in\mathcal{D}} \max(-\rho(\signal, \Always_{[aT, bT]} (s > 0)), 0) + \gamma (a - b)
\label{eq:pSTL problem}
\vspace{-2mm}
\end{equation}
}
where $\gamma$ is a coefficient on the term $a-b$ that encourages the interval to be larger.
We solve \eqref{eq:pSTL problem} via gradient descent, using the smooth time interval mask discussed in Sec.~\ref{sec: differentiable time}.
Specifically, we annealed the time interval mask scaling factor and the temperature with a sigmoid schedule, and performed 5000 gradient steps with a step size of $10^{-2}$. To ensure $0 \leq a < b \leq 1$, we passed them through a sigmoid function first.
Fig.~\ref{fig:time interval gradient descent} shows a few snapshots of the gradient steps and the loss landscape as the scaling parameter and temperature increase. We can observe that our solution converges to the global optimum and is consistent with the ground truth values (subject to the injected noise).
To give a sense of the computation time, using JAX on a M2 MacbookPro, it took about
$122\pm 0.479\mu$s per gradient step.

\section{Conclusion}
We present \stlcgpp, a masking-based approach for computing STL robustness using automatic differentiation libraries. \stlcgpp\ mimics the operations that underpin transformer architectures, and we demonstrate the computational, theoretical, and practical benefits of the proposed masking approach over \stlcg, which uses a recurrent approach.
The observed advantages of masking over recurrent operations mirror the advantages of using transformers over recurrent neural networks for processing sequential data.
We also present two \stlcgpp\ libraries in JAX and PyTorch, demonstrating their usage in several robotics-related problems such as machine learning, trajectory planning, and deep generative modeling.
\stlcgpp\ offers significant computational advantages over \stlcg, thus presenting new and exciting opportunities for incorporating STL specifications into various online robot planning and control tasks that require fast computation and inference speeds.

\bibliographystyle{IEEEtran}
\bibliography{../../../../bib/main,../../../../bib/ctrl_papers}

\newcommand{\noopsort}[1]{} \newcommand{\printfirst}[2]{#1} \newcommand{\singleletter}[1]{#1} \newcommand{\switchargs}[2]{#2#1}
\begin{thebibliography}{10}
\providecommand{\url}[1]{#1}
\csname url@samestyle\endcsname
\providecommand{\newblock}{\relax}
\providecommand{\bibinfo}[2]{#2}
\providecommand{\BIBentrySTDinterwordspacing}{\spaceskip=0pt\relax}
\providecommand{\BIBentryALTinterwordstretchfactor}{4}
\providecommand{\BIBentryALTinterwordspacing}{\spaceskip=\fontdimen2\font plus
\BIBentryALTinterwordstretchfactor\fontdimen3\font minus \fontdimen4\font\relax}
\providecommand{\BIBforeignlanguage}[2]{{%
\expandafter\ifx\csname l@#1\endcsname\relax
\typeout{** WARNING: IEEEtran.bst: No hyphenation pattern has been}%
\typeout{** loaded for the language `#1'. Using the pattern for}%
\typeout{** the default language instead.}%
\else
\language=\csname l@#1\endcsname
\fi
#2}}
\providecommand{\BIBdecl}{\relax}
\BIBdecl

\bibitem{MalerNickovic2004}
O.~Maler and D.~Nickovic, ``{Monitoring Temporal Properties of Continuous Signals},'' \emph{{Lecture Notes in Computer Science}}, no. 3253, pp. 152--166, 2004.

\bibitem{PantAbbasEtAl2017}
Y.~V. Pant, H.~Abbas, and R.~Mangharam, ``{Smooth Operator: Control using the Smooth Robustness of Temporal Logic},'' in \emph{{IEEE Conf.\ on Control Technology and Applications}}, 2017.

\bibitem{MengFan2024}
Y.~Meng and C.~Fan, ``{Diverse Controllable Diffusion Policy With Signal Temporal Logic},'' \emph{{IEEE Robotics and Automation Letters}}, vol.~9, no.~10, pp. 8354--8361, 2024.

\bibitem{LiuMehdipourEtAl2021}
W.~Liu, N.~Mehdipour, and C.~Belta, ``{Recurrent Neural Network Controllers for Signal Temporal Logic Specifications Subject to Safety Constraints},'' \emph{{IEEE Control Systems Letters}}, vol.~6, pp. 91 -- 96, 2021.

\bibitem{LeungArechigaEtAl2021}
K.~Leung, N.~Ar\'{e}chiga, and M.~Pavone, ``{Backpropagation through signal temporal logic specifications: Infusing logical structure into gradient-based methods},'' \emph{{Int.\ Journal of Robotics Research}}, 2022.

\bibitem{ZhongRempeEtAl2023}
Z.~Zhong, D.~Rempe, D.~Xu, Y.~Chen, S.~Veer, T.~Che, B.~Ray, and M.~Pavone, ``{Guided Conditional Diffusion for Controllable Traffic Simulation},'' in \emph{{Proc.\ IEEE Conf.\ on Robotics and Automation}}, 2023.

\bibitem{LeungPavone2022}
K.~Leung and M.~Pavone, ``{Semi-Supervised Trajectory-Feedback Controller Synthesis for Signal Temporal Logic Specifications},'' in \emph{{American Control Conference}}, 2022.

\bibitem{VeerLeungEtAl2023}
S.~Veer, K.~Leung, R.~Cosner, Y.~Chen, and M.~Pavone, ``{Receding Horizon Planning with Rule Hierarchies for Autonomous Vehicles},'' in \emph{{Proc.\ IEEE Conf.\ on Robotics and Automation}}, 2023.

\bibitem{DeCastroLeungEtAl2020}
J.~DeCastro, K.~Leung, N.~Ar\'{e}chiga, and M.~Pavone, ``{Interpretable Policies from Formally-Specified Temporal Properties},'' in \emph{{Proc.\ IEEE Int.\ Conf.\ on Intelligent Transportation Systems}}, 2020.

\bibitem{HochreiterSchmidhuber1997}
S.~Hochreiter and J.~Schmidhuber, ``{Long short-term memory},'' \emph{{Neural Computation}}, 1997.

\bibitem{VaswaniShazeerEtAl2017}
A.~Vaswani, N.~Shazeer, N.~Parmar, J.~Uszkoreit, L.~Jones, A.~N. Gomez, L.~Kaiser, and I.~Polosukhin, ``{Attention is All You Need},'' in \emph{{Conf.\ on Neural Information Processing Systems}}, 2017.

\bibitem{DiamondBoyd2016}
S.~Diamond and S.~Boyd, ``{{CVXPY}: A {P}ython-Embedded Modeling Language for Convex Optimization},'' \emph{{Journal of Machine Learning Research}}, vol.~17, no.~83, pp. 1--5, 2016.

\bibitem{Tedrakeothers2019}
R.~Tedrake \emph{et~al.}, ``{Drake: Model-based design and verification for robotics},'' \emph{{Available at } \url{https://drake.mit.edu}}, 2019.

\bibitem{YamaguchiHoxhaEtAl2023}
T.~Yamaguchi, B.~Hoxha, and D.~Nickovic, ``{{RTAMT}: Runtime Robustness Monitors with Application to CPS and Robotics},'' \emph{{Int.\ Journal on Software Tools for Technology Transfer}}, vol.~26, pp. 79--99, 2023.

\bibitem{Balakrishnan2021}
A.~Balakrishnan, ``{{Signal Temporal Logic C++ Toolbox}},'' \emph{{Available at } \url{https://github.com/anand-bala/signal-temporal-logic}}, 2021.

\bibitem{KurtzLin2022}
V.~Kurtz and H.~Lin, ``{Mixed-Integer Programming for Signal Temporal Logic with Fewer Binary Variables},'' \emph{{IEEE Control Systems Letters}}, vol.~6, pp. 2635--2640, 2022.

\bibitem{CardonaLeahyEtAl2023}
G.~A. Cardona, K.~Leahy, M.~Mann, and C.-I. Vasile, ``{A Flexible and Efficient Temporal LogicTool for Python: PyTeLo},'' \emph{{Available at }\url{https://arxiv.org/abs/2310.08714}}, 2023.

\bibitem{VazquezChanlatte2022}
M.~Vazquez-Chanlatte, ``{{pySTLToolbox}},'' \emph{{Available at }\url{https://github.com/mvcisback/py-signal-temporal-logic}}, 2022.

\bibitem{BartocciDeshmukhEtAl2018}
E.~Bartocci, J.~Deshmukh, A.~Donz{\'{e}}, G.~Fainekos, O.~Maler, D.~Nickovic, and S.~Sankaranarayanan, ``{Specification-based Monitoring of Cyber-Physical Systems: A Survey on Theory, Tools and Applications},'' \emph{{Lectures on Runtime Verification}}, 2018.

\bibitem{MehdipourVasileEtAl2019}
N.~Mehdipour, C.-I. Vasile, and C.~Belta, ``{Arithmetic-Geometric Mean Robustness for Control from Signal Temporal Logic Specifications},'' in \emph{{American Control Conference}}, 2019.

\bibitem{HeBartocciEtAl2022}
J.~He, E.~Bartocci, D.~Ničković, H.~Iakovic, and R.~Grosu, ``{{DeepSTL} -- From English Requirements to Signal Temporal Logic},'' in \emph{{IEEE/ACM Int.\ Conf.\ on Software Engineering}}, 2022.

\bibitem{JAX2018}
J.~Bradbury, R.~Frostig, P.~Hawkins, M.~J. Johnson, C.~Leary, D.~Maclaurin, G.~Necula, A.~Paszke, J.~Vander{P}las, S.~Wanderman-{M}ilne, and Q.~Zhang, ``{{JAX}: composable transformations of {P}ython+{N}um{P}y programs},'' \emph{Available at \url{http://github.com/google/jax}}, 2018.

\bibitem{PaszkeGrossEtAl2017}
A.~Paszke, S.~Gross, S.~Chintala, G.~Chanan, E.~Yang, Z.~DeVito, Z.~Lin, A.~Desmaison, L.~Antiga, and A.~Lerer, ``{Automatic differentiation in {PyTorch}},'' in \emph{{Conf.\ on Neural Information Processing Systems - Autodiff Workshop}}, 2017.

\bibitem{SongMengEtAl2021}
J.~Song, C.~Meng, and S.~Ermon, ``{Denoising Diffusion Implicit Models},'' in \emph{{Int.\ Conf.\ on Learning Representations}}, 2021.

\bibitem{JannerDuEtAl2022}
M.~Janner, Y.~Du, J.~Tenenbaum, and S.~Levine, ``{Planning with Diffusion for Flexible Behavior Synthesis},'' in \emph{{Int.\ Conf.\ on Machine Learning}}, 2022.

\bibitem{MizutaLeung2024}
K.~Mizuta and K.~Leung, ``{{CoBL-Diffusion}: Diffusion-Based Conditional Robot Planning in Dynamic Environments Using Control Barrier and Lyapunov Functions},'' in \emph{{IEEE/RSJ Int.\ Conf.\ on Intelligent Robots \& Systems}}, 2024.

\bibitem{KapoorKangEtAl2024}
P.~Kapoor, E.~Kang, and R.~Meira-G{\'o}es, ``Safe planning through incremental decomposition of signal temporal logic specifications,'' in \emph{{NASA Formal Methods Symposium}}, 2024.

\bibitem{BartocciManjunathEtAl2021}
E.~Bartocci, N.~Manjunath, L.~Mariani, C.~Mateis, and D.~Ni\v{c}kovi\'{c}, ``{{CPSDebug}: Automatic failure explanation in {CPS} models},'' \emph{{Int.\ Journal on Software Tools for Technology Transfer}}, 2021.

\bibitem{KangGanlathEtAl2024}
E.~Kang, A.~Ganlath, S.~Mishra, F.~Baiduc, and N.~Ammar, ``{Contract-Driven Runtime Adaptation},'' in \emph{{NASA Formal Methods Symposium}}, 2024.

\end{thebibliography}

\end{document}